\documentclass{article}

\PassOptionsToPackage{numbers, sort&compress}{natbib}
\usepackage[preprint]{neurips_2026}

\usepackage[utf8]{inputenc}
\usepackage[T1]{fontenc}
\usepackage{hyperref}
\usepackage{url}
\usepackage{booktabs}
\usepackage{amsfonts}
\usepackage{fancyvrb}
\usepackage{amsmath}
\usepackage{graphicx}
\usepackage{xcolor}
\usepackage{pifont} 
\usepackage{booktabs}
\usepackage{tikz}
\usetikzlibrary{arrows.meta, positioning, calc, backgrounds, fit}

\title{OpenKedge: Governing Agentic Mutation with Execution-Bound Safety and Evidence Chains}

\author{
	Jun He \\
	OpenKedge.io \\
	\texttt{junhe@openkedge.io} \\
	\and 
	Deying Yu \\
	OpenKedge.io \\
	\texttt{deying@openkedge.io}
}

\begin{document}
	
	\maketitle
	
\begin{abstract}
The rise of autonomous AI agents exposes a fundamental flaw in API-centric architectures: probabilistic systems directly execute state mutations without sufficient context, coordination, or safety guarantees. We introduce OpenKedge, a protocol that redefines mutation as a governed process rather than an immediate consequence of API invocation.
OpenKedge requires actors to submit declarative intent proposals, which are evaluated against deterministically derived system state, temporal signals, and policy constraints prior to execution. Approved intents are compiled into execution contracts that strictly bound permitted actions, resource scope, and time, and are enforced via ephemeral, task-oriented identities. This shifts safety from reactive filtering to preventative, execution-bound enforcement.
Crucially, OpenKedge introduces an Intent-to-Execution Evidence Chain (IEEC), which cryptographically links intent, context, policy decisions, execution bounds, and outcomes into a unified lineage. This transforms mutation into a verifiable and reconstructable process, enabling deterministic auditability and reasoning about system behavior.
We evaluate OpenKedge across multi-agent conflict scenarios and cloud infrastructure mutations. Results show that the protocol deterministically arbitrates competing intents and cages unsafe execution while maintaining high throughput, establishing a principled foundation for safely operating agentic systems at scale.
\end{abstract}

\section{Introduction}

Modern software systems are increasingly operated by autonomous AI agents \cite{wang2024agentsurvey, openai2025agents}. While enabling unprecedented automation across APIs, databases, and cloud infrastructure, this shift introduces profound risks: incorrect or context-unaware agent actions can trigger service outages and cascading failures. As the industry rapidly moves toward AI-orchestrated workflows, ensuring the safety of agentic operations is an existential challenge.

However, this shift exposes a fundamental mismatch: modern infrastructure relies on \emph{passive APIs} that blindly execute mutations upon invocation. APIs inherently assume callers are deterministic, correct, and isolated—assumptions that fail entirely in probabilistic, agentic environments. Traditional systems enforce syntactic correctness and access control, but completely ignore semantic intent, hidden dependencies, and multi-actor coordination. Consequently, seemingly valid requests can easily trigger cascading outages \cite{aws2025outage, azure2024outage, crowdstrike2024outage}. As AI agents are increasingly deployed in production, they amplify both the speed and scale of these potential failures \cite{openai2025agents, anthropic2024agents}.

This gap is particularly glaring in agent-driven DevOps. AI systems often propose infrastructure changes without complete visibility into active cross-service dependencies or real-time traffic. Thus, a perfectly formatted, fully authorized API call can still be catastrophic. Recognizing this requires a fundamental architectural pivot: from merely \emph{executing} mutations to \emph{governing} them based on intent, global context, and enforceable constraints.

As agents transition from sandboxes to production, the flaws of API-centric architectures compound. Recent evaluations reveal systemic issues: API hallucinations, conflicting collaborative actions, and unsafe mutations driven by stale context \cite{xi2025agentbench, anthropic2024agents, openai2025agents}. Crucially, these failures are not just symptoms of immature LLM reasoning—they represent a deep architectural deficit. By treating mutation as an immediate execution obligation rather than a governed decision process, current systems leave a dangerous loophole: authorized actions are not necessarily safe actions.

These flaws cause concrete operational failures. If a scheduling agent blindly flags a service as \texttt{offline} based on outdated data while an operator marks it \texttt{online}, traditional APIs accept both, creating an oscillating, corrupted state. In cloud environments, an agent might confidently delete an ``unused'' database, oblivious to hidden secondary workloads. Improving an LLM's reasoning is insufficient; the underlying system itself must actively evaluate, coordinate, and constrain mutations before execution.

To fix this, we introduce \textbf{OpenKedge}, an intent-based protocol for governing state mutation in agentic systems. Rather than exposing direct, executable APIs, OpenKedge requires agents to submit structured intent proposals. The system then rigorously evaluates these proposals against real-time global context and strict policy constraints before any execution is allowed.

We identify four fundamental challenges in deploying agents over passive APIs:
\begin{itemize}
	\item \textbf{Lack of Contextual Validation:} API requests are executed in isolation, ignoring recent updates, temporal constraints, and system-wide invariants.
	\item \textbf{Multi-Agent Conflict:} Multiple actors (human and AI) issue concurrent, mutually exclusive updates without deterministic resolution.
	\item \textbf{Unsafe Mutation:} Probabilistic agents may generate hallucinated or contextually incorrect actions, leading to harmful or irreversible state changes.
	\item \textbf{Absence of End-to-End Traceability:} Traditional architectures lack comprehensive end-to-end logging, making it impossible to reason about a mutation action initialized by AI agents. Compared to change management tools for manual changes, these agent-driven API actions are significantly less formalized, lacking mechanisms for review, visibility, and post-incident reasoning.
\end{itemize}

OpenKedge addresses these failures by establishing \emph{intent proposals} as the standard interaction unit. Instead of issuing commands, agents declare desired outcomes, which are evaluated against global context and system policy. This governance extends to execution: approved proposals are translated into explicit execution contracts bound by resource scopes and temporal limits. These contracts are enforced through short-lived, task-oriented identities. This guarantees that probabilistically generated intents, once approved, are executed with deterministic, bounded authority. 

Underlying OpenKedge is a fact-based, event-sourced representation of system truth. This enables deterministic state derivation and complete auditability. To manage concurrent multi-actor updates, an embedded coordination layer automatically resolves conflicts using actor authority, trust scores, and temporal signals. 

Every phase of this lifecycle is preserved in an Intent-to-Execution Evidence Chain (IEEC), which serves as an append-only log. By recording the actor (\emph{who}), timestamp (\emph{when}), intent (\emph{what}), policy (\emph{how}), context, and outcome, the IEEC ensures that every decision and execution step is traceable and reconstructible.

We validate OpenKedge across two distinct domains: a reference implementation governing real-time operational state for physical businesses, and a high-impact cloud infrastructure simulation managing DevOps agent mutations. These evaluations demonstrate that conflicting proposals can be deterministically arbitrated, and destructive automation loops safely caged at the execution boundary, without degrading system scale or throughput.

Our primary contributions are threefold:
\begin{itemize}
\item \textbf{Intent-Governed Mutation Protocol.} We formulate OpenKedge, an intent-based governance protocol that restructures state mutation in agentic systems. By decoupling intent from execution and evaluating proposals against global context and policy, the system ensures that only semantically valid and conflict-free mutations are admitted.

\item \textbf{Execution-Bound Safety via Contracts.} We introduce execution contracts enforced through ephemeral, task-oriented identities that strictly bound permitted actions, resource scope, and temporal validity. This guarantees that even under hallucinated or adversarial agent behavior, physical execution remains provably constrained.

\item \textbf{Intent-to-Execution Evidence Chain (IEEC).} We propose IEEC, a cryptographically linked lineage that binds intent, context, policy decisions, execution bounds, and outcomes into a unified event structure. IEEC establishes a new invariant for agentic systems: every mutation must be both execution-bounded and explainable in lineage, enabling deterministic auditability and reasoning over system behavior.
\end{itemize}

Together, these components drive a paradigm shift: from passive execution APIs to active, intent-driven governance. OpenKedge establishes a core safety guarantee: system state evolves exclusively through valid, bounded, and conflict-free transitions derived from approved intent.

\section{Related Work}

The emergence of autonomous agents capable of issuing actions across software systems has motivated a range of approaches for controlling and safeguarding agent behavior. Existing work primarily focuses on enabling tool use, enforcing access control, or validating actions at runtime. While these approaches address important aspects of safety, they largely assume that mutations are executed directly through APIs and therefore operate as auxiliary safeguards rather than redefining the mutation model itself.

In contrast, OpenKedge introduces a fundamentally different abstraction: mutation is no longer treated as an immediate consequence of API invocation, but as a governed process mediated by intent, context, policy, and execution constraints. This shift requires rethinking not only how actions are validated, but how they are represented, coordinated, and enforced throughout their lifecycle.
\subsection{AI Safety, Runtime Control, and Mutation Governance}

Foundational AI safety research has long identified the severe risks associated with autonomous systems executing consequential real-world actions~\cite{amodei2016concrete}. As a result, there is a growing operational focus on runtime safety paradigms that shift the safety boundary from model outputs to action execution.

Recent frameworks such as the Autonomous Action Runtime Management (AARM) specification advocate intercepting and evaluating AI-driven actions at runtime, emphasizing control over tool invocation and execution~\cite{errico2026aarm}. Similarly, practical agent execution frameworks, such as Anthropic's Claude Code, employ client-side interception techniques—such as regex-based filtering and abstract syntax tree (AST) validation—to detect potentially dangerous commands~\cite{claudecode2025}. These approaches represent an important step toward mitigating unsafe actions in agentic systems.

However, these mechanisms remain fundamentally reactive and execution-centric. They operate on already-formed actions, treating safety as a filtering or interception problem, and lack access to system-wide context, resource semantics, or multi-agent coordination. As a result, they cannot reason about whether an action is appropriate within the broader system state, nor can they provide guarantees about consistency or conflict resolution across concurrent actors.

OpenKedge builds on the intuition that safety must extend beyond runtime interception, but takes a fundamentally different approach by redefining mutation itself as a governed process. Rather than validating actions after they are generated, OpenKedge requires all mutations to originate as structured intent proposals, which are evaluated against system-wide context and policy before execution. Approved intents are translated into execution contracts and enforced through short-lived, task-oriented identities, ensuring that execution remains strictly bounded and aligned with the approved scope.

This shift moves the problem from controlling \emph{how actions are executed} to governing \emph{what state transitions are allowed and how they are safely realized}. In doing so, OpenKedge unifies context-aware reasoning, policy enforcement, multi-agent coordination, and execution-bound safety into a single mutation lifecycle, providing guarantees that extend beyond the capabilities of traditional safety or runtime control mechanisms. OpenKedge thus complements existing safety mechanisms while addressing a broader systems problem: ensuring that mutations are not only permitted, but correct, coordinated, and safely executed.

\subsection{LLM Agents and Tool Use}

A significant body of recent work has focused on enabling Large Language Models (LLMs) to interact seamlessly with external tools and APIs \cite{yao2023react, schick2023toolformer}. While these frameworks substantially enhance an agent's reasoning and action-planning capabilities, they largely perpetuate the reliance on direct, ungoverned API invocation. Recent comprehensive evaluations have underscored the compounding reliability and coordination challenges inherent in such direct-invocation systems \cite{wang2024agentsurvey, xi2025agentbench}. OpenKedge directly complements this research by shifting the focus from agent capability to system reliability, introducing a systemic abstraction that dictates how autonomous actions are safely validated and applied. 

\subsection{Multi-Agent Systems}

Traditional research in LLM-based multi-agent systems primarily addresses the complexities of coordination, communication protocols, and distributed decision-making among autonomous actors. Recent frameworks such as AutoGen \cite{wu2023autogen}, MetaGPT \cite{hong2023metagpt}, and ChatDev \cite{qian2023chatdev} have established sophisticated paradigms for role specialization, conversational programming, and simulated workflows. However, the majority of this literature focuses on optimizing agent behavior, prompt engineering, and intra-agent reasoning rather than architecting safe, systemic mutation pathways for shared state. OpenKedge addresses this critical gap by providing a deterministic coordination layer that programmatically resolves conflicting state updates utilizing explicitly defined parameters of authority, trust, and temporal constraints.

\subsection{Trust and Reliability in Multi-Agent Systems}

The formalization of trust has long been a foundational challenge in traditional multi-agent architectures. Classic computational trust models were developed to quantitatively evaluate agent reliability based on historical interactions and partner selection \cite{sabater2005review}. With the widespread adoption of LLM-based autonomous agents, inter-agent trust has gained renewed importance due to the inherently probabilistic reasoning pathways of large language models. 

Recent studies highlight a ``trust paradox'' in LLM environments: while escalating inter-agent trust accelerates collaborative workflows, it simultaneously expands the attack surface for instruction laundering, hallucination cascades, and over-authorization \cite{xu2025trustparadox}. To mitigate these risks, contemporary decentralized frameworks evaluate dynamic trust using execution heuristics to safeguard underlying communication protocols \cite{decentralizedtrust2025}. 

OpenKedge advances this domain by transitioning trust from a peer-to-peer communicative filter into a structural arbitration mechanism. By mathematically integrating dynamic trust scores—specifically alongside static role authority and temporal recency—directly into the governance engine, OpenKedge systematically arbitrates concurrent state conflicts. Consequently, the architecture predicts and resolves authority races, ensuring that highly trusted, authoritative intents deterministically override unverified or hallucinated proposals prior to execution.

\subsection{API-Based Systems}

Modern software architectures predominantly rely on APIs as the primary interface for state mutation. This paradigm inherently assumes callers operate deterministically, possess sufficient context, and construct valid requests—delegating correctness entirely to the client. Recent literature on Agent-Computer Interfaces (ACIs) highlights the severe limitations of this model, demonstrating that traditional APIs, designed as rigid machine-to-machine contracts, are notoriously brittle and inadequate when exposed to probabilistically reasoning agents~\cite{yang2024sweagent}. When this abstraction inevitably breaks down in agentic environments, it leads to unsafe or nonsensical mutations. OpenKedge departs from the API-centric model entirely. Instead of relying on client-side correctness, it introduces an explicit governance layer that evaluates proposed mutations against system-wide constraints before any execution occurs.

\subsection{Event Sourcing}

Event sourcing represents system state as an append-only sequence of immutable events, ensuring rigorous auditability and reproducibility \cite{fowler2005event}. OpenKedge builds upon this foundational paradigm by integrating structured facts with explicit temporal validity and by strictly governing which events are permitted to mutate the system state. Unlike traditional event-sourced architectures that generally assume the validity of incoming events, OpenKedge treats all incoming actions as untrusted proposals, enforcing validation through a mandatory policy layer.

The challenge of safely merging concurrent, distributed state updates is well-studied in distributed systems theory, most notably through Convergent and Commutative Replicated Data Types (CRDTs) \cite{shapiro2011comprehensive}. CRDTs provide rigorous mathematical guarantees for resolving concurrent mutations without centralized locking or distributed consensus protocols. While OpenKedge's fact-based state composition shares the CRDT goal of deterministic conflict resolution, it diverges by explicitly incorporating semantic policy, hierarchical actor authority, and temporal validity into the reduction function. This tailors the resolution process specifically to govern the unpredictable and potentially adversarial nature of autonomous agents.

\subsection{Summary}

While prior work successfully addresses isolated aspects of representation, coordination, and safety, the literature lacks a unified abstraction for governing how shared state evolves under the influence of autonomous, agentic systems. OpenKedge bridges this critical gap by synthesizing intent-based mutation, policy-driven validation, and fact-based state representation into a single, cohesive architectural protocol for agent-driven operations. We further establish an end-to-end mutation correctness property: all reachable system states are composed exclusively of valid, non-conflicting, and execution-bounded transitions derived from policy-approved intent. This guarantee is grounded in the Intent-to-Execution Evidence Chain (IEEC), which enforces that every mutation is both execution-bounded and accompanied by a verifiable, causally linked decision lineage.

\section{Problem Definition and System Overview}

\subsection{Failure Modes in Agentic Systems}

We identify five fundamental failure modes that arise when agent-driven mutations are executed through passive APIs without contextual governance:

\begin{itemize}
	\item \textbf{Contextual Blindness:} API requests are executed in isolation, ignoring recent updates, temporal constraints, and system-wide domain invariants.
	\item \textbf{Multi-Agent Conflict:} Multiple actors concurrently issue mutually exclusive updates without a deterministic resolution mechanism.
	\item \textbf{Unsafe Mutation:} Probabilistic agents acting under uncertainty may generate hallucinated or invalid actions, causing harmful state changes.
	\item \textbf{Temporal Inconsistency:} Decisions based on stale system views result in actions that are valid locally but incorrect globally.
	\item \textbf{Unbounded Authority:} Execution relies on broad, persistent credentials, allowing mutations to exceed their intended scope.
\end{itemize}

\begin{figure}[t]
\centering
\resizebox{\textwidth}{!}{ 
\begin{tikzpicture}[
    node distance=0.8cm and 0.6cm,
    base/.style={draw, rectangle, rounded corners=3pt, align=center, minimum height=1.1cm, text width=2.4cm, font=\small, thick},
    api/.style={base, fill=gray!10, draw=gray!70!black},
    danger/.style={base, fill=red!5, draw=red!70!black},
    ok/.style={base, fill=blue!5, draw=blue!70!black},
    safe/.style={base, fill=green!5, draw=green!70!black},
    arrow/.style={->, >=Stealth, thick, draw=black!70},
    bullet_text/.style={align=left, font=\footnotesize, text width=12cm}
]

\node[font=\bfseries, anchor=north west] (title1) at (0, 0) {Traditional API-Centric Mutation};

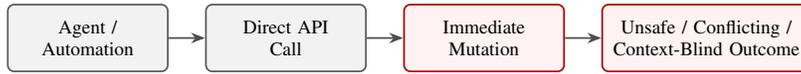
\node[api] (a1) [below=0.4cm of title1.south west, anchor=north west] {Agent /\\ Automation};
\node[api] (a2) [right=of a1] {Direct API\\ Call};
\node[danger] (a3) [right=of a2] {Immediate\\ Mutation};
\node[danger, text width=3.2cm] (a4) [right=of a3] {Unsafe / Conflicting /\\ Context-Blind Outcome};

\draw[arrow] (a1) -- (a2);
\draw[arrow] (a2) -- (a3);
\draw[arrow] (a3) -- (a4);

\node[bullet_text] (bullets1) [below=0.4cm of a1.south west, anchor=north west] {
    $\bullet$ Caller is implicitly trusted \\
    $\bullet$ No semantic validation \\
    $\bullet$ No system-wide context check \\
    $\bullet$ No bounded execution authority
};

\draw[dashed, draw=gray!40, thick] ([xshift=-0.2cm, yshift=-0.8cm]bullets1.south west) -- +(14.8, 0);

\node[font=\bfseries, anchor=north west] (title2) [below=1.6cm of bullets1.south west, anchor=north west] {OpenKedge-Governed Mutation};

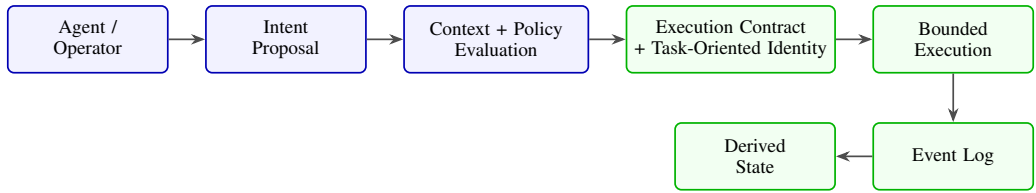
\node[ok] (b1) [below=0.4cm of title2.south west, anchor=north west] {Agent /\\ Operator};
\node[ok] (b2) [right=of b1] {Intent\\ Proposal};
\node[ok, text width=2.8cm] (b3) [right=of b2] {Context + Policy\\ Evaluation};
\node[safe, text width=3.2cm] (b4) [right=of b3] {Execution Contract\\ + Task-Oriented Identity};
\node[safe] (b5) [right=of b4] {Bounded\\ Execution};

\node[safe] (b6) [below=of b5] {Event Log};
\node[safe] (b7) [left=of b6] {Derived\\ State};

\draw[arrow] (b1) -- (b2);
\draw[arrow] (b2) -- (b3);
\draw[arrow] (b3) -- (b4);
\draw[arrow] (b4) -- (b5);
\draw[arrow] (b5) -- (b6); 
\draw[arrow] (b6) -- (b7); 

\node[bullet_text] (bullets2) [below=0.4cm of b1.south west |- b7.south, anchor=north west] {
    $\bullet$ Mutation begins from intent, not raw invocation \\
    $\bullet$ Context and policy are evaluated before execution \\
    $\bullet$ Authority is bounded by a short-lived task-oriented identity \\
    $\bullet$ Effects are recorded and state is derived deterministically
};

\end{tikzpicture}
}
\caption{
Comparison between traditional API-centric mutation and OpenKedge-governed mutation. In conventional systems, agents or automation issue direct API calls, and mutations are executed immediately under the assumption that the caller is correct, context-aware, and non-conflicting. This often leads to unsafe, conflicting, or context-blind outcomes. OpenKedge replaces direct invocation with an intent-based pipeline in which proposed mutations are evaluated against context and policy, translated into constrained execution contracts, and enforced through short-lived, task-oriented identities before execution. The resulting effects are recorded as events from which system state is deterministically derived.
}
\label{fig:failure-vs-openkedge}
\end{figure}

These failures stem from a structural flaw in API-centric execution: the implicit trust placed in callers to provide correct, context-aware actions. Correctness cannot be guaranteed at the client or API boundary; it must be enforced through a centralized governance layer. Figure~\ref{fig:failure-vs-openkedge} contrasts this traditional assumption with OpenKedge's intent-based pipeline, which structurally decouples desired outcomes from execution authority.

\subsection{System Model}

OpenKedge formalizes system mutation as a governed transformation from an untrusted proposal to a deterministic state change. We define the following core entities and the principles that govern them:

\paragraph{Actors and Intent.}
The system comprises diverse actors (autonomous agents, automation, and humans). Instead of executing direct API calls, actors must declare desired outcomes through \emph{intent proposals}. This principle of \emph{intent-first mutation} decouples the \emph{what} from the \emph{how}, allowing the system to neutrally evaluate the request before determining execution mechanics.

\paragraph{Context and Policy.}
Every intent is evaluated against current system context—including recent state, dependencies, and temporal variables. Decision-making is governed by deterministic policies that encode authorization, conflict-resolution logic, and operational constraints. This \emph{policy-driven governance} ensures all mutations are contextually validated and reproducible.

\paragraph{Contracts and Bounded Execution.}
Approved intents are translated into explicit \emph{execution contracts} bounding the permitted action, target resource, and temporal validity. To prevent privilege escalation, these contracts are enforced by dynamically generated, short-lived \emph{task-oriented identities}. This \emph{identity-enforced safety} guarantees that execution authority inherently cannot exceed its approved scope.

\paragraph{Events and State Derivation.}
All mutations are recorded as an append-only stream of facts. System state is deterministically derived from this log. This \emph{deterministic state derivation} provides complete auditability and temporal reasoning, ensuring that every transition in the system is fully traceable.

Table~\ref{tab:openkedge_comparison} situates these properties against existing agent frameworks and runtime safeguards, illustrating OpenKedge's unique integration of multi-agent safety and bounds-constrained execution.

\begin{table}[t]
	\centering
	\small
	\newcommand{\cmark}{\ding{51}}
	\newcommand{\xmark}{\ding{55}}
	\begin{tabular}{lcccc}
		\toprule
		\textbf{System} & \textbf{Direct Mutation} & \textbf{Context-Aware} & \textbf{Multi-Agent Safe} & \textbf{Governed Mutation} \\
		\midrule
		API-based Systems & \cmark & \xmark & \xmark & \xmark \\
		Workflow Engines & \cmark & Partial & \xmark & \xmark \\
		ReAct Agents~\cite{yao2023react} & \cmark & Partial & \xmark & \xmark \\
		Toolformer~\cite{schick2023toolformer} & \cmark & \xmark & \xmark & \xmark \\
		Agent Frameworks~\cite{wang2024agentsurvey, xi2025agentbench} & \cmark & Partial & \xmark & \xmark \\
		AARM~\cite{errico2026aarm} & Partial & Partial & \xmark & Partial \\
		Claude Code~\cite{claudecode2025} & \cmark & \xmark & \xmark & Partial \\
		\midrule
		\textbf{OpenKedge (Ours)} & \xmark & \cmark & \cmark & \cmark \\
		\bottomrule
	\end{tabular}
	\caption{
	Comparison of OpenKedge with existing approaches. Prior systems primarily rely on direct mutation or runtime filtering, with limited context awareness and no built-in support for multi-agent safety. AARM introduces runtime action control, and Claude Code applies heuristic validation (e.g., regex- and AST-based checks), but both lack comprehensive context-aware governance. In contrast, OpenKedge introduces intent-based, policy-governed mutation with execution-bound enforcement and native multi-agent coordination.
	}
	\label{tab:openkedge_comparison}
\end{table}

\subsection{The OpenKedge Architecture}

\begin{figure}[t]
\centering
\resizebox{0.95\textwidth}{!}{
\begin{tikzpicture}[
    >=Stealth,
    font=\small,
    node distance=0.7cm,
    module/.style={draw, rectangle, rounded corners=3pt, align=center, minimum width=4.6cm, minimum height=0.8cm, thick},
    input/.style={module, fill=gray!10},
    proposal/.style={module, fill=blue!10, draw=blue!70!black},
    gov/.style={module, fill=orange!10, draw=orange!70!black},
    exec/.style={module, fill=green!10, draw=green!70!black},
    bus/.style={draw=violet!70!black, fill=violet!5, thick, rectangle, rounded corners=3pt, align=center, minimum width=1.0cm, minimum height=10.5cm},
    state/.style={module, fill=teal!10, draw=teal!70!black, minimum width=3.6cm},
    arr/.style={->, thick},
    logarr/.style={->, dashed, violet!70!black},
    residual/.style={->, thick, black!40, rounded corners=6pt}
]

\node[input] (agent) {Agent / Operator};
\node[proposal] (intent) [above=of agent] {Intent Proposal};

\node[gov] (context) [above=1.6cm of intent] {Context Expansion};
\node[gov] (policy) [above=0.8cm of context] {Policy Evaluation};

\node[exec] (contract) [above=1.4cm of policy] {Execution Contract};
\node[exec] (identity) [above=0.8cm of contract] {Task-Oriented Identity};
\node[exec] (execution) [above=0.8cm of identity] {Execution};

\node[input] (sys) [above=1.4cm of execution] {Target Systems / APIs};

\draw[arr] (agent) -- (intent);
\draw[arr] (intent) -- (context);
\draw[arr] (context) -- (policy);
\draw[arr] (policy) -- (contract);
\draw[arr] (contract) -- (identity);
\draw[arr] (identity) -- (execution);
\draw[arr] (execution) -- (sys);

\begin{scope}[on background layer]
    \node[draw=orange!40, fill=orange!2, dashed, thick, rounded corners=5pt,
          fit=(context) (policy), inner sep=10pt] (govbox) {};
    \node[anchor=south west, xshift=-0.3cm, yshift=0.15cm, font=\small\bfseries, color=orange!70!black] 
        at (govbox.north west) {Governance Layer};

    \node[draw=green!40, fill=green!2, dashed, thick, rounded corners=5pt,
          fit=(contract) (execution), inner sep=10pt] (execbox) {};
    \node[anchor=south west, xshift=-0.3cm, yshift=0.15cm, font=\small\bfseries, color=green!70!black] 
        at (execbox.north west) {Enforcement Layer};
\end{scope}

\node[bus, right=6.0cm of contract, yshift=-1.0cm] (log) {\rotatebox{270}{\textbf{Evidence Chain (IEEC)}}};
\node[state, below=1.2cm of log] (state) {Derived State Store};

\draw[arr] (log.south) -- node[right, font=\footnotesize] {projects} (state.north);

\draw[arr]
(state.west) -- ++(-4.0,0) |- node[pos=0.4, right, align=left, font=\footnotesize] {
\textbf{Context Signals}\\
- Resource Status \\
- Dependency Maps \\
- Historical Aggregates \\
- Quotas
} (context.east);

\foreach \node/\label in {
    intent/actor \& action,
    policy/policy trace,
    contract/bounds,
    execution/results \& time
} {
    \draw[logarr]
        (\node.east) -- (log.west |- \node.east)
        node[midway, above, font=\footnotesize] {\label};
}

\draw[residual]
(intent.west) -- ++(-1.0,0) |- node[pos=0.25, left, font=\footnotesize] {Intent}
(contract.west);

\draw[residual]
(contract.west) -- ++(-0.8,0) |- node[pos=0.25, left, font=\footnotesize] {Bounds}
(execution.west);

\end{tikzpicture}
}
\caption{
OpenKedge architecture with IEEC as a cross-cutting system backbone. Mutations originate as intent proposals and are evaluated through governance (context and policy) before being translated into execution contracts. These contracts are enforced via short-lived, task-oriented identities, ensuring that execution remains strictly bounded. All actions are recorded in an Intent-to-Execution Evidence Chain (IEEC), from which system state is deterministically derived and fed back into future decisions.
}
\label{fig:openkedge-architecture}
\end{figure}
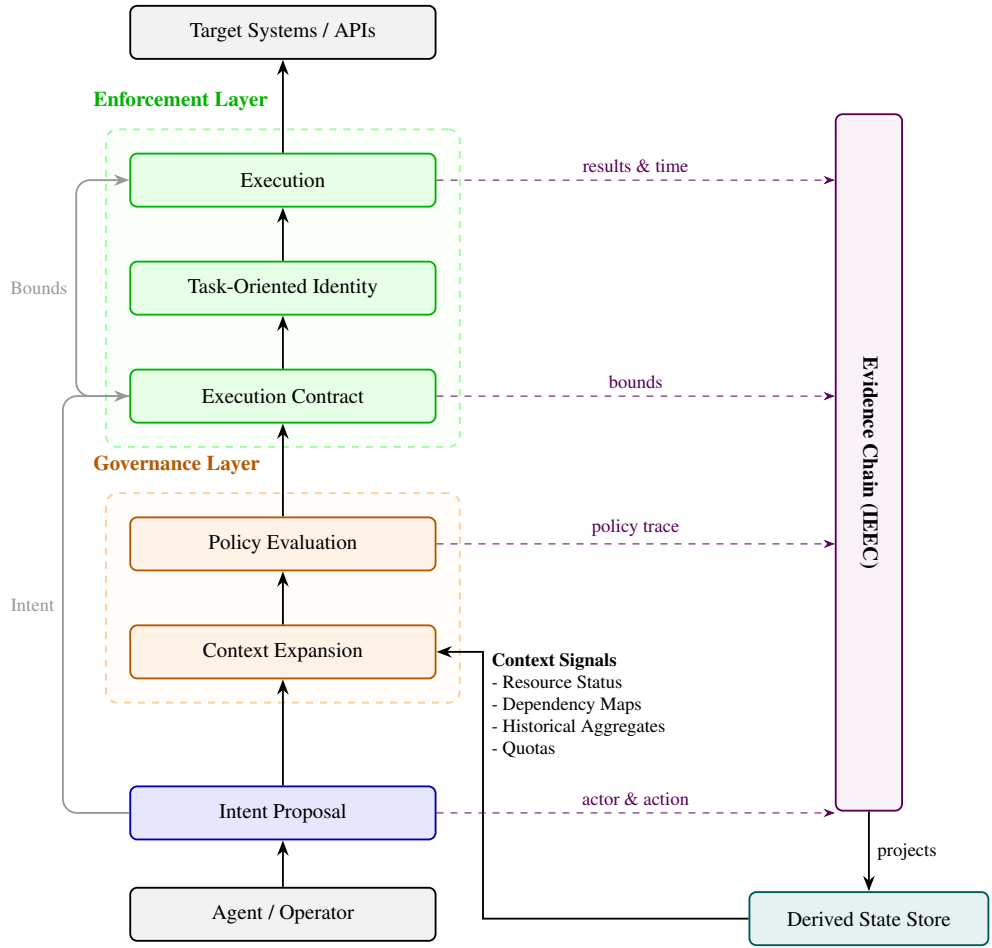

OpenKedge replaces direct API execution with a structured pipeline that actively mediates all mutations through interpretation, validation, and enforcement. As illustrated in Figure~\ref{fig:openkedge-architecture}, the architecture transforms intent proposals into bounded execution, recording the outcomes as immutable events from which all system state is deterministically derived.

\paragraph{1. Intent Formation and Contextualization.}
Mutations begin as structured intent proposals. To inform decision-making, OpenKedge gathers semantic context from the active system state, enabling reasoning that extends beyond static API definitions.

\paragraph{2. Policy Evaluation and Contract Generation.}
The enriched proposal is evaluated against deterministic policy rules. Upon approval, OpenKedge generates an explicit \emph{execution contract}---formally a tuple $C = (a, r, t)$ denoting the permitted action $a$, resource scope $r$, and temporal validity $t$. This contract serves as the binding interface bridging decision and execution.

\paragraph{3. Identity-Bound Execution.}
The execution contract is physically enforced by a short-lived, task-oriented identity scoped strictly to the contractual boundaries. The mutation is executed against the target infrastructure using this identity, neutralizing the risk of unauthorized privilege expansion.

\paragraph{4. Intent-to-Execution Evidence Chain (IEEC).}
To provide end-to-end explainability, OpenKedge records every phase of this lifecycle in an Intent-to-Execution Evidence Chain (IEEC). Unlike conventional audit logs that capture passive API invocations, the IEEC cryptographically links the initial proposal, retrieved context, policy decision, execution bounds, and final outcome. For instance, an agent shutting down a virtual machine yields a comprehensive trace encompassing its identity, policy justification, and dynamic bounds. Consequently, any state change can be deterministically reconstructed to answer not just \emph{what} occurred, but \emph{why} and \emph{how} it was authorized.

\subsection{Threat Model and Trust Assumptions}

We define the safety boundaries of OpenKedge to precisely scope its mitigation guarantees. The core governance pipeline (Context, Policy Engine, IEEC, and State Derivation) operates within a trusted computing base backed by reliable infrastructure. Conversely, all proactive input sources—including human operators, verified agents, and unverified components—are treated as strictly untrusted.

\paragraph{In-Scope Threats.}
OpenKedge structurally mitigates the following systemic and probabilistic failures:
\begin{itemize}
	\item \textbf{Semantic Hallucinations:} Syntactically valid but contextually destructive actions generated by probabilistic reasoning errors.
	\item \textbf{Temporal Asynchrony:} ``Stale-state'' mutations generating distributed race conditions.
	\item \textbf{Multi-Agent Collisions:} Uncoordinated, concurrent operations targeting shared entities.
	\item \textbf{Authority Override:} Unauthorized attempts by autonomous agents to circumvent semantic constraints or human operational authority.
\end{itemize}

\paragraph{Out-of-Scope Threats.}
The protocol does not protect against the cryptographic compromise of the underlying event log, direct administrative infrastructure tampering, or adversarial cognitive attacks (e.g., prompt injections or model jailbreaks). OpenKedge explicitly assumes that cognitive attacks will successfully hijack agent reasoning; rather than attempting to safeguard internal model logic, the architecture mitigates the resulting damage via zero-trust containment at the execution boundary.

\section{Formal Truth Representation}

In agentic environments, multiple autonomous actors concurrently execute mutations under partial and asynchronous contexts. Maintaining a globally consistent system state under such volatility demands a representation that is observable and causal. Traditional architectures relying on in-place state mutation inherently obscure decision pathways, limiting the system's capacity to reason about multi-agent coordination.

To resolve this, OpenKedge adopts an event-sourced architecture. Rather than treating instantaneous state as the primary computational object, system truth is exclusively derived from an append-only sequence of immutable events. By modeling state as a pure function of mutation history, OpenKedge guarantees deterministic state reconstruction and facilitates causal reasoning across temporal horizons.

\subsection{Cryptographic Event Representation and Evidence Lineage}

Every phase of the mutation lifecycle—intent proposal, contextual expansion, policy evaluation, contract derivation, and execution—is materialized as a discrete event appended to an immutable log.
Formally, a lifecycle event $e$ is defined as the tuple:
\[
e = (i, c, t, r)
\]
where $i$ encodes the originating intent, $c$ the environmental context utilized during evaluation, $t$ a monotonically increasing timestamp, and $r$ the resulting state delta or execution outcome.

These events are causally linked via cryptographic hashes to form the Intent-to-Execution Evidence Chain (IEEC), an unbroken, strictly ordered chain capturing the full provenance of every mutation.

\subsection{Functional State Derivation}

System state is never directly overwritten. Instead, it is derived as a fold over the event history. Let $E = \langle e_1, e_2, \dots, e_n \rangle$ denote the globally ordered event sequence. The system state $S$ at index $n$ is defined by:
\[
S_n = f(E_{\le n})
\]
where $f$ denotes a deterministic state projection function.

This functional architecture enforces three critical properties: \emph{Determinism}, guaranteeing an identical event sequence yields an identical state; \emph{Auditability}, ensuring every state transition is cryptographically traceable back to its originating intent; and \emph{Replayability}, enabling historical state reconstruction for forensic debugging, verification, or agent training.

\subsection{Causal Substrate for Multi-Agent Coordination}

In multi-agent topologies, the immutable event log acts as the shared substrate for coordination. By decomposing macroscopic mutations into causally linked event sequences, the governance layer can mathematically reason about the causal relationships between disparate actions, including data dependencies, ordering constraints, and semantic resource conflicts.

Operating over event history rather than volatile state natively enables the detection of interacting mutations across uncoordinated agents and drives the deterministic resolution of competing intent proposals. Grounding policy decisions in an unforgeable event history ensures context-aware reasoning safely scales across multiple actors and execution windows.

\subsection{Summary}

This functional event representation serves as the bedrock for OpenKedge's execution paradigms. While intent evaluation and contract enforcement guarantee mutations are authorized at runtime, the event log provides the complementary guarantee that these effects remain observable, reproducible, and causally sound.

\section{Execution Model and Contract Enforcement}

While the previous section formalizes system truth as a deterministic derivation of event history, the architecture must equally guarantee that the physical execution mechanics precisely reflect the approved intent. OpenKedge achieves this by translating policy decisions into explicit execution contracts enforced by task-oriented identities, structurally locking physical actuation to the exact dimensions of the governance decision.

\subsection{Execution as an Event Generator}

Rather than treating execution as an opaque, terminal side-effect of API dispatch, OpenKedge mathematically models execution as an event-producing transducer. Every authorized physical mutation generates a verifiable cryptographic trace reflecting its runtime materialization. Because the event log rejects any outcome failing to map back to a verified contract, OpenKedge physically entangles operational side-effects directly into the causal event fabric. Consequently, the Intent-to-Execution Evidence Chain (IEEC) is not an auxiliary logging mechanism, but a structural invariant of the protocol: execution is permitted if and only if it produces a verifiable, causally linked lineage entry.

\subsection{Execution Contracts and Identity Enforcement}

Following policy approval, declarative intent is compiled into an explicit execution contract $C = (a, R, T)$, where $a$ restricts the permitted action, $R$ defines the immutable target resource bounding, and $T$ enforces strict temporal expiry. 

To enforce these contracts without modifying underlying orchestration services, OpenKedge provisions \emph{task-oriented identities}. A task-oriented identity is an ephemeral credential dynamically synthesized for a single execution sequence. By forcing execution engines to assume identities intrinsically locked to $C$ parameters—rather than relying on broad service accounts—OpenKedge physically isolates the agent environment. Malicious instructions, hijacked control flows, and probabilistic hallucinations are mathematically constrained from violating the contract boundary, completely bounding the blast radius of any individual agent failure.

\subsection{Formal System Invariants}

We formalize the correctness of the OpenKedge execution model through three invariants connecting intent, execution, and system state.

Let $E$ denote the append-only event log, where each event $e_k \in E$ is produced by physical execution associated with contract $C_k$. Let $\mathcal{C}$ denote the set of all valid execution contracts derived from policy, and $S = f(E)$ denote the state folded deterministically from the event log.

\paragraph{Invariant 1: Execution-Event Consistency.}
Every event $e_k \in E$ must correspond to an execution contract $C_k \in \mathcal{C}$ such that $e_k \in \text{Exec}(C_k)$. Formally, $E \subseteq \bigcup_{C \in \mathcal{C}} \text{Exec}(C)$. This invariant guarantees \textbf{State Safety}: all reachable system states are composed strictly of authorized, contract-bounded transitions.

\paragraph{Invariant 2: Multi-Agent Conflict Safety.}
Let $\text{conflict}(C_i, C_j)$ denote incompatible proposed state transitions targeting overlapping dimensions. For any pair where $\text{conflict}(C_i, C_j) = \text{true}$, at most one of the corresponding executions may produce events:
\[
\neg \exists \, e_i \in \text{Exec}(C_i), \; e_j \in \text{Exec}(C_j) \quad \text{such that both } e_i, e_j \in E.
\]
This ensures derived state is structurally decoupled from execution interleaving or arbitrary network race conditions.

\paragraph{Invariant 3: Liveness and Progress.}
Assuming a fair execution environment, every admissible (non-conflicting, highest-priority) contract $C \in \mathcal{C}$ is mathematically guaranteed to execute and produce at least one event $e \in \text{Exec}(C)$ within the system log. No valid mutation is indefinitely starved.

\paragraph{Theorem: End-to-End Mutation Correctness.}
Unifying these invariants, we prove that derived state $S = f(E)$ is composed exclusively of valid, non-conflicting, and eventually realized transitions. By structurally treating execution as an event-producing transducer governed by task-oriented identities, OpenKedge ensures that mutation correctness is preserved completely at the atomic level of system history.
\section{Multi-Agent Coordination and Policy Governance}

A defining characteristic of agentic environments is the concurrent operation of heterogeneous actors—including human operators, autonomous agents, and legacy systems—over shared resources. Because these actors frequently propose state transitions under partial and potentially conflicting contexts, robust coordination is paramount. OpenKedge introduces a deterministic governance framework to arbitrate concurrent operations, resolving conflicting updates natively at the intent layer prior to physical execution.

\subsection{Semantic Conflict Detection and Composition}

Let $A = \{a_1, a_2, \dots, a_m\}$ denote a set of actors issuing a concurrent batch of intent proposals $P = \{p_1, p_2, \dots, p_k\}$. Through contextual evaluation, each proposal $p_i$ is mapped to a candidate mutation defined by a set of asserted propositional facts $F_i$ over the system state.

We define a symmetric Boolean conflict relation:
\[
\text{Conflict}(F_i, F_j) \in \{\text{true}, \text{false}\}
\]

A semantic conflict emerges if and only if two proposals assert mutually exclusive constraints or induce incompatible state transitions over intersecting resource domains. Proposals operating over disjoint resources or asserting strictly orthogonal constraints evaluate to $\text{false}$ and are jointly admissible.

\subsection{Authority, Trust, and Temporal Arbitration}

When $\text{Conflict}(F_i, F_j) = \text{true}$, OpenKedge arbitrates the contention using a multidimensional resolution function encompassing static authority, dynamic trust, and temporal recency. Let each actor $a_i \in A$ possess a discrete role-based $\text{Authority}(a_i)$ and a continuously updated performance metric $\text{Trust}(a_i) \in [0,1]$.

The protocol derives a scalar priority score for each competing intent proposal $p_i$ authored by $a_i$:
\[
\text{Priority}(p_i) = \alpha \cdot \text{Authority}(a_i) + \beta \cdot \text{Trust}(a_i)
\]
where $\alpha$ and $\beta$ are system-defined weighting coefficients determining the relative influence of organizational hierarchy versus empirical reliability.

To prevent temporal inversion and handle asynchronous propagation delays gracefully, the resolution function incorporates strict temporal bounds relative to the current system clock $t_{\text{now}}$:
\[
\text{Recency}(p_i) = t_{\text{now}} - t_{\text{origin}}(p_i)
\]

By evaluating $\text{Priority}(p_i)$ subject to a maximum allowable $\text{Recency}(p_i)$ threshold, the policy engine mathematically guarantees that stale, low-trust, or unauthorized probabilistic actions cannot preempt recent, highly-authoritative state transitions. In edge cases resulting in irreducible ambiguity (e.g., precise priority scalar parity between concurrent agents), OpenKedge escalates the decision rather than committing an unsafe mutation.

\subsection{Policy-as-Code Evaluation and Traceability}

The Policy Engine serves as the ultimate governance arbiter. Conflict resolution operates as a pure function of the current policy, system context, and immutable event history $E$:
\[
\text{Evaluate}(p_i, \text{Context}, E) \in \{\text{Approve}, \text{Reject}, \text{Escalate}\}
\]

To align with modern infrastructural best practices, policies are constructed under the \emph{policy-as-code} paradigm~\cite{openpolicyagent, peebles2024cedar}. Decoupling governance logic from underlying application code enables transparently version-controlled and deterministically verifiable multi-agent administration. 

Furthermore, every governance evaluation is materialized as a cryptographic tuple $(\text{Intent}, \text{Context}, \text{Decision}, \text{EvaluatedRules})$ within the Intent-to-Execution Evidence Chain (IEEC). This structurally transforms the historically opaque process of multi-agent conflict resolution into a mathematically verifiable decision trace. By lifting conflict resolution entirely to the algorithmic intent layer, OpenKedge eradicates the volatile network race conditions that plague uncoordinated distributed agents.
\section{Implementation}

To empirically demonstrate the feasibility and performance of the OpenKedge protocol, we implemented a reference system called \textbf{Riftront}. Designed to govern the operational state of a multi-actor platform, Riftront serves as a minimal, representative instantiation of the intent-based pipeline.

\subsection{Riftront Reference Architecture and Coordination}

Riftront realizes the OpenKedge protocol through a decoupled, lock-free architecture. A centralized \emph{Ingress Interface} normalizes intent proposals originating from diverse actors---including AI agents, automation scripts, and human operators. Because system state is dynamically folded from an immutable event log rather than managed via in-place database updates, Riftront completely bypasses distributed entity locking. This allows the centralized \emph{Policy Engine} to arbitrate multi-dimensional conflicts---such as competing proposals for a shared resource---using deterministic temporal and trust constraints natively at the evaluation layer.

\subsection{Policy Engine Integration using Cedar}

To implement deterministic policy-as-code, the prototype integrates the \textbf{Cedar} policy language~\cite{peebles2024cedar}, which provides formal verification capabilities and sub-millisecond evaluation latency.

Each incoming intent is translated into a native Cedar evaluation request formatted as the relational tuple $(\text{Principal}, \text{Action}, \text{Resource}, \text{Context})$. The \emph{Context} object securely injects the continuously derived system state alongside critical temporal signals and computed trust heuristics. This formulation enables declarative governance that natively arbitrates multi-agent priority.

The core governance of Riftront is demonstrated through localized temporal protection rules that prioritize human authority over autonomous agents. For example, a policy \emph{forbidding} agent-driven state updates if a human operator has modified the entity within the previous 3600 seconds (one hour) and the agent's trust score remains below a threshold of 0.8 can be expressed cleanly in Cedar syntax:

\begin{center}
\begin{tikzpicture}
\node[draw=gray!40, fill=gray!5, thick, rounded corners=4pt, inner sep=12pt, align=left, text width=0.85\textwidth, font=\small\ttfamily] {
\textcolor{blue!70!black}{\textbf{forbid}} (\\
\hspace*{1.5em}principal \textcolor{blue!70!black}{\textbf{in}} Role::\textcolor{orange!60!black}{"Agent"},\\
\hspace*{1.5em}action == Action::\textcolor{orange!60!black}{"UpdateOperatingStatus"},\\
\hspace*{1.5em}resource\\
)\\
\textcolor{blue!70!black}{\textbf{when}} \{\\
\hspace*{1.5em}context.time\_since\_owner\_update < \textcolor{violet!70!black}{3600} \&\&\\
\hspace*{1.5em}context.trust\_score < \textcolor{violet!70!black}{0.8}\\
\};
};
\end{tikzpicture}
\end{center}

Cedar yields a binary \texttt{Allow} or \texttt{Deny} decision, which the architecture maps directly to \texttt{Approve} or \texttt{Reject}, while rules annotated for human-in-the-loop workflows result in deterministic \texttt{Escalate} outcomes.

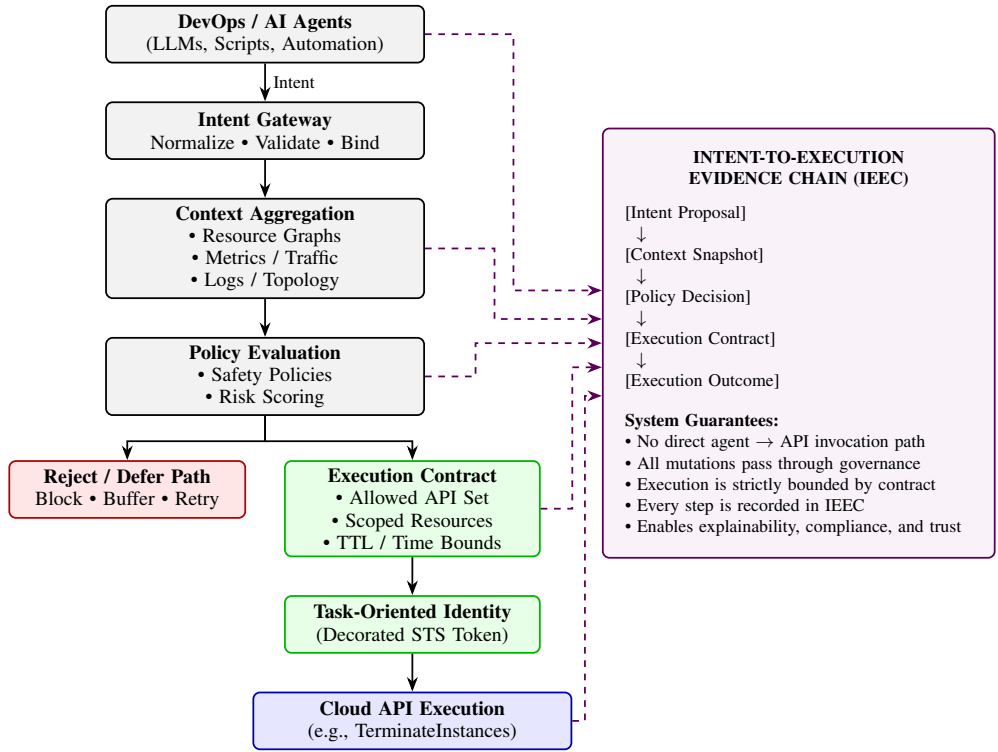
\begin{figure}[htbp]
\centering
\resizebox{0.95\textwidth}{!}{
\begin{tikzpicture}[
    >=Stealth,
    node distance=0.6cm and 0.8cm,
    box/.style={draw, rectangle, rounded corners=3pt, align=center, thick, fill=gray!10, text width=4.8cm, minimum height=0.9cm, font=\small},
    reject/.style={box, fill=red!10, draw=red!70!black, text width=3.5cm},
    accept/.style={box, fill=green!10, draw=green!70!black, text width=3.8cm},
    ieec/.style={draw=violet!70!black, fill=violet!5, thick, rectangle, rounded corners=3pt, align=left, text width=5.5cm, font=\footnotesize, inner sep=10pt},
    arr/.style={->, thick},
    residual/.style={->, thick, dashed, violet!70!black}
]

\node[box] (agent) {\textbf{DevOps / AI Agents}\\(LLMs, Scripts, Automation)};
\node[box, below=0.6cm of agent] (gateway) {\textbf{Intent Gateway}\\Normalize • Validate • Bind};
\node[box, below=0.6cm of gateway] (context) {\textbf{Context Aggregation}\\• Resource Graphs\\• Metrics / Traffic\\• Logs / Topology};
\node[box, below=0.6cm of context] (policy) {\textbf{Policy Evaluation}\\• Safety Policies\\• Risk Scoring};

\path (policy.south) -- ++(0,-0.4) coordinate (fork);
\node[reject, left=0.3cm of fork, anchor=north east, yshift=-0.3cm] (reject) {\textbf{Reject / Defer Path}\\Block • Buffer • Retry};
\node[accept, right=0.3cm of fork, anchor=north west, yshift=-0.3cm] (contract) {\textbf{Execution Contract}\\• Allowed API Set\\• Scoped Resources\\• TTL / Time Bounds};

\node[accept, below=0.6cm of contract] (identity) {\textbf{Task-Oriented Identity}\\(Decorated STS Token)};
\node[box, fill=blue!10, draw=blue!70!black, below=0.6cm of identity] (execute) {\textbf{Cloud API Execution}\\(e.g., TerminateInstances)};

\draw[arr] (agent) -- node[right, font=\footnotesize] {Intent} (gateway);
\draw[arr] (gateway) -- (context);
\draw[arr] (context) -- (policy);
\draw[arr] (policy.south) -- (fork) -| (reject.north);
\draw[arr] (policy.south) -- (fork) -| (contract.north);
\draw[arr] (contract) -- (identity);
\draw[arr] (identity) -- (execute);

\node[ieec, right=2.8cm of context.east, anchor=west, yshift=-1.5cm] (evidence) {
    {\centering \textbf{INTENT-TO-EXECUTION\\EVIDENCE CHAIN (IEEC)}\par}
    \vspace{0.2cm}
    {[Intent Proposal]} \\
    {~~$\downarrow$} \\
    {[Context Snapshot]} \\
    {~~$\downarrow$} \\
    {[Policy Decision]} \\
    {~~$\downarrow$} \\
    {[Execution Contract]} \\
    {~~$\downarrow$} \\
    {[Execution Outcome]} \\[0.3cm]
    \textbf{System Guarantees:}\\
    • No direct agent $\rightarrow$ API invocation path\\
    • All mutations pass through governance\\
    • Execution is strictly bounded by contract\\
    • Every step is recorded in IEEC\\
    • Enables explainability, compliance, and trust
};

\draw[residual] (agent.east) -- ++(1.4,0) |- (evidence.165);
\draw[residual] (context.east) -- ++(1.1,0) |- (evidence.173);
\draw[residual] (policy.east) -- ++(0.8,0) |- (evidence.180);
\draw[residual] (contract.east) -- ++(0.5,0) |- (evidence.187);
\draw[residual] (execute.east) -- ++(0.2,0) |- (evidence.195);

\end{tikzpicture}
}
\caption{OpenKedge DevOps workflow and Intent-to-Execution Evidence Chain (IEEC). Agents submit intent proposals instead of executing APIs directly. The pipeline evaluates context and policy before compiling approved intents into execution-bounded contracts enforced via task-oriented identities. Every stage of the mutation lifecycle—from intent formation to execution outcome—is recorded within the IEEC as a cryptographically linked lineage. This ensures that execution is permitted if and only if it produces a verifiable decision trace, transforming infrastructure mutation into a deterministic, auditable, and replayable process.}
\label{fig:devops-workflow}
\end{figure}

\subsection{Cloud Infrastructure Implementation and Traceability}

To validate the protocol beyond application-level state, we developed an execution adapter for Amazon Web Services (AWS)---a domain where autonomous agent hallucinations can trigger catastrophic service outages. In typical cloud environments, agents operate using long-lived service accounts with broad permissions. A hallucinating agent issuing an \texttt{ec2:TerminateInstances} command (a request to delete compute resources) is often only limited by its static permissions, not the current system context.

OpenKedge intercepts these direct API calls and converts them into intent proposals. These proposals are evaluated against the current state of the infrastructure, retrieved via topology graphs and real-time traffic identifiers (utilizing \emph{AWS Config} and \emph{Elastic Load Balancing}). 

If an intent is approved, the protocol generates a strictly bounded \emph{execution contract}. This contract is enforced systemically through the synthesis of short-lived cryptographic credentials via the \emph{AWS Security Token Service (STS)}. Rather than utilizing a static service account, the agent is issued an ephemeral IAM session policy tailored specifically to the approved contract. By assuming this \emph{task-oriented identity}—where permissions are dynamically scoped strictly to the precise \emph{Amazon Resource Name (ARN)} of the target entity—the agent is cryptographically prevented from accessing any resource outside its mandate, even if its internal logic fails or enters a destructive loop.

Throughout this lifecycle, the system materializes the decision lineage into an Intent-to-Execution Evidence Chain (IEEC), implemented via AWS CloudWatch Logs. Unlike traditional audit systems such as \emph{AWS CloudTrail}, which record isolated API invocations post hoc, the IEEC captures the full causal lineage of mutation decisions—including intent formation, contextual signals, policy evaluation traces, execution contracts, and final outcomes—into a unified, cryptographically linked record. This elevates auditability from passive observation to deterministic verification: every mutation is not only recorded, but can be reconstructed as a proof of why it was permitted and how it was safely executed. Consequently, infrastructure mutations transition from opaque operational actions into fully explainable and replayable decision artifacts. This directly realizes the invariant introduced in Section~5: execution is permitted if and only if it produces a verifiable lineage within the IEEC, ensuring that every mutation is both execution-bounded and explainable in lineage. The resulting DevOps workflow is illustrated in Figure~\ref{fig:devops-workflow}.

\section{Evaluation}

We evaluate OpenKedge empirically across three dimensions critical to autonomous systems: conflict arbitration correctness, execution safety under hallucination, and system determinism. Representative scenarios capturing primary failure modes were simulated against the Riftront architecture.

\subsection{Conflict Arbitration and Correctness}

We subjected a high-concurrency ecosystem combining human operators, verified agents, and unverified agents to hostile workloads. This included diametrically opposed state proposals, extensive concurrent modifications, and systematically stale context inputs.

\paragraph{Authority Conflicts.} When a human operator proposed an \texttt{open} state concurrently with an agent proposing a \texttt{closed} state, the governance engine correctly evaluated the static authority differential, deterministically rejecting the agent's intent. 

\paragraph{Temporal and Trust Contention.} When equal-authority agents raced to asynchronously overwrite identical operational metrics, OpenKedge isolated and executed the most recent, highest-trust intent entirely without relying on distributed execution locks. Orthogonal intents concurrently targeting separate partitions naturally merged into the folded log seamlessly.

\subsection{High-Impact Infrastructure Safety}

To assess safety within high-stakes domains, we evaluated OpenKedge against simulated AI-driven outages on shared cloud infrastructure.

\paragraph{Scenario 1: Unsafe Resource Deletion.}
An agent proposed terminating a compute instance that its internal model probabilistically inferred to be ``unused.'' Although the proposed API call was syntactically well-formed, the instance actually supported active downstream dependencies. By evaluating the intent against the live AWS Config resource topology, the policy engine preemptively rejected the context-blind mutation prior to any state-mutating API dispatch.

\paragraph{Scenario 2: Traffic-Blind Scaling.}
An agent proposed a syntactically well-formed cluster capacity reduction. However, the proposal violated system availability constraints given the peak active load-balancing traffic. The evaluation layer intercepted this anomaly and blocked the execution, successfully isolating the context-blind mutation.

Across all simulated infrastructure workloads, unsafe state transitions were consistently and predictably caged prior to execution. Furthermore, admitted mutations operated strictly within the confines of ephemeral, scope-bound STS tokens, deterministically validating the protocol's capacity for strict blast-radius containment. In all scenarios, IEEC enabled deterministic reconstruction of decision pathways, allowing post-hoc validation of both accepted and rejected mutations.

\subsection{Determinism and Performance}

To evaluate algorithmic determinism, we subjected the pipeline to a concurrent workload of 10,000 asynchronous proposals across multiple actors targeting shared entities. Submitting identical execution workloads produced strictly identical execution contracts, event logs, and derived system states across sequential runs, demonstrating the systematic prevention of observable asynchronous race conditions.

Overhead profiling was executed within an AWS environment utilizing an Amazon EC2 \texttt{m5.2xlarge} instance (8 vCPUs, 32 GiB RAM) for the core evaluation engine. Compute was backed by Intel Xeon Platinum processors (up to 3.1 GHz) running on the AWS Nitro System hypervisor, which systematically limits traditional virtualization jitter. The decision lineage and state materialization were persisted to a managed Amazon RDS PostgreSQL event store, deployed on a dedicated \texttt{db.m5.2xlarge} instance. To ensure storage bandwidth did not bottleneck the write-ahead log (WAL) during high-velocity state transitions, the database was backed by \texttt{gp3} Elastic Block Store (EBS) volumes provisioned with a baseline of 6,000 IOPS.

Under this sustained load, policy evaluation averaged ~11 ms per request. By actively pruning historical evaluation context arrays, the state derivation sequence maintained a 99th percentile latency of < 30 ms. Leveraging the lock-free formulation and predictable storage I/O, the architecture sustained operations peaking at ~3,200 mutations per second without observable throughput degradation, confirming its viability for massive-scale enterprise integration.
\section{Discussion}

\subsection{Architectural Paradigm Shift}

OpenKedge represents a fundamental shift in system design: moving from the \emph{execution} of mutation requests to the \emph{governance} of them. Traditional systems entangle intent, decision, and state, treating mutation as an immediate side effect of API invocation. This model assumes that callers are correct, context-aware, and deterministic---assumptions that fail in agentic environments.

OpenKedge explicitly separates what an agent wants to achieve (Intent), whether the system allows it (Decision), how the mutation is physically constrained (Execution Contract), and the resulting ground truth (State). As probabilistic agents increasingly replace deterministic operators, explicit mutation governance becomes a structurally necessary foundation rather than an optional abstraction.

\subsection{Execution-Bound Safety as a First-Class Primitive}

A central insight of this work is that safety in agentic systems must extend beyond decision-time validation to include execution-time enforcement. Traditional systems rely on authorization mechanisms that validate whether an action is permitted, but they do not guarantee that execution adheres strictly to the intended scope.

By enforcing approved proposals through explicitly scoped, cryptographic execution contracts rather than static API keys, OpenKedge guarantees that execution logic remains strictly bounded. Even if upstream components like agent reasoning models hallucinate, the downstream execution layer prevents unauthorized side effects. Therefore, OpenKedge shifts system safety from a heuristic best-effort property into a guaranteed physical property of execution. OpenKedge introduces a new invariant for agentic systems: every mutation must be both bounded in execution and explainable in lineage.

\subsection{Generalizability and Limitations}

While the reference implementation evaluates OpenKedge within operation heuristics and AWS cloud infrastructure, the protocol generalizes to any domain managing concurrent, multi-actor mutation under partial information, including IoT networks and distributed ledgers.

A current limitation of the architecture is its reliance on manually authored deterministic policies (e.g., Cedar rules). While declarative rules provide rigid safety, they require explicit human foresight. Future work will explore learning-based policy adaptation and tighter integration with formal verification engines, allowing the governance layer to dynamically infer safety invariants from historic event topologies.

\section{Conclusion}

The rapid integration of autonomous AI agents marks a fundamental shift in how software systems are operated. As probabilistic agents increasingly generate and execute actions across distributed infrastructure, legacy assumptions of caller correctness, complete context, and safe execution no longer hold. Passive, API-centric architectures inherently expose systems to unsafe and conflicting mutations.

OpenKedge addresses this challenge by redefining mutation as a governed process. By requiring declarative intent proposals, evaluating them against global context and policy, and enforcing execution through bounded contracts and task-oriented identities, the system ensures that all mutations are both semantically valid and physically constrained.

Crucially, the Intent-to-Execution Evidence Chain (IEEC) binds every mutation to a verifiable decision lineage, establishing a new invariant for agentic systems: execution is permitted if and only if it is both bounded in action and explainable in lineage. This transforms mutation from an opaque operational side effect into a deterministic, auditable, and reconstructable process.

As AI agents become primary operators of digital infrastructure, the correctness of individual agents becomes secondary to the correctness of the system governing them. OpenKedge provides a principled foundation for this shift, unifying intent governance, execution-bound safety, and verifiable mutation lineage into a coherent protocol for safely operating agentic systems at scale.

\section*{Code Availability}

The OpenKedge protocol and reference implementation are available at:
\url{https://github.com/openkedge/openkedge}.

	\bibliographystyle{unsrtnat}
	\bibliography{references}

@article{fowler2005event,
	title = {Event Sourcing},
	author = {Fowler, Martin},
	year = {2005},
	note = {\url{https://martinfowler.com/eaaDev/EventSourcing.html}},
}

@article{yao2023react,
	title = {ReAct: Synergizing Reasoning and Acting in Language Models},
	author = {Yao, Shunyu and others},
	journal = {arXiv preprint arXiv:2210.03629},
	year = {2023},
}

@article{wang2024agentsurvey,
	title = {A Survey on Large Language Model based Autonomous Agents},
	author = {Wang, Lei and others},
	journal = {arXiv preprint arXiv:2308.11432},
	year = {2024},
}

@article{openai2025agents,
	title = {Practices for Building Reliable Agents},
	author = {OpenAI},
	journal = {Technical Report},
	year = {2025},
}

@article{anthropic2024agents,
	title = {On the Safety and Reliability of AI Agents},
	author = {Anthropic},
	journal = {Technical Report},
	year = {2024},
}

@article{yang2024sweagent,
	title = {SWE-agent: Agent-Computer Interfaces Enable Automated Software
	         Engineering},
	author = {Yang, John and Jimenez, Carlos E. and Wettig, Alexander and Luan,
	          Kilian and Lin, Shunyu and Narasimhan, Karthik and Yao, Shunyu},
	journal = {arXiv preprint arXiv:2405.15793},
	year = {2024},
}

@article{xi2025agentbench,
	title = {AgentBench: Evaluating LLM-based Agents},
	author = {Xi, Zhiheng and others},
	journal = {arXiv preprint arXiv:2308.03688},
	year = {2025},
}

@article{schick2023toolformer,
	title = {Toolformer: Language Models Can Teach Themselves to Use Tools},
	author = {Schick, Timo and others},
	journal = {arXiv preprint arXiv:2302.04761},
	year = {2023},
}

@article{sabater2005review,
	title = {Review on computational trust and reputation models},
	author = {Sabater, Jordi and Sierra, Carles},
	journal = {Artificial Intelligence Review},
	volume = {24},
	number = {1},
	pages = {33--60},
	year = {2005},
	publisher = {Springer},
}

@misc{openpolicyagent,
	title = {Open Policy Agent},
	author = {{Styra, Inc.}},
	howpublished = {\url{https://www.openpolicyagent.org}},
	year = {2023},
}

@inproceedings{peebles2024cedar,
	author = {Peebles, Craig and others},
	title = {Cedar: A New Language for Expressive and Fast Authorization},
	booktitle = {Proceedings of the 18th USENIX Symposium on Operating Systems
	             Design and Implementation (OSDI '24)},
	year = {2024},
	publisher = {USENIX Association},
}

@article{xu2025trustparadox,
	title = {The Trust Paradox in LLM-Based Multi-Agent Systems: When
	         Collaboration Becomes a Security Vulnerability},
	author = {Xu, Zijie and others},
	journal = {arXiv preprint arXiv:2510.18563},
	year = {2025},
}

@article{decentralizedtrust2025,
	title = {Decentralized Multi-Agent System with Trust-Aware Communication},
	author = {Anonymous},
	journal = {arXiv preprint arXiv:2512.02410},
	year = {2025},
}

@article{shapiro2011comprehensive,
	title = {A comprehensive study of Convergent and Commutative Replicated Data
	         Types},
	author = {Shapiro, Marc and Pregui{\c{c}}a, Nuno and Baquero, Carlos and
	          Zawirski, Marek},
	journal = {Inria Research Report},
	year = {2011},
}

@article{wu2023autogen,
	title = {AutoGen: Enabling Next-Gen LLM Applications via Multi-Agent
	         Conversation},
	author = {Wu, Qingyun and Bansal, Gagan and Zhang, Jieyu and Wu, Yiran and
	          Li, Beibin and Zhu, Erkang and Jiang, Li and Zhang, Xiaoyun and
	          Zhang, Shaokun and Liu, Jiale and others},
	journal = {arXiv preprint arXiv:2308.08155},
	year = {2023},
}

@article{hong2023metagpt,
	title = {MetaGPT: Meta Programming for A Multi-Agent Collaborative Framework
	         },
	author = {Hong, Sirui and Zhuge, Mingchen and Chen, Jonathan and Zheng,
	          Xiawu and Cheng, Yuheng and Zhang, Ceyao and Wang, Jinlin and Wang,
	          Zili and Yau, Steven K. and Lin, Zijian and others},
	journal = {arXiv preprint arXiv:2308.00352},
	year = {2023},
}

@article{qian2023chatdev,
	title = {ChatDev: Communicative Agents for Software Development},
	author = {Qian, Chen and Cong, Xin and Yang, Cheng and Chen, Weize and Su,
	          Yusheng and Xu, Juyuan and Liu, Zhiyuan and Ma, Zhiyuan},
	journal = {arXiv preprint arXiv:2307.07924},
	year = {2023},
}

@article{errico2026aarm,
	title = {Autonomous Action Runtime Management (AARM): A System Specification
	         for Securing AI-Driven Actions at Runtime},
	author = {Errico, Herman},
	journal = {arXiv preprint arXiv:2602.09433},
	year = {2026},
}

@article{amodei2016concrete,
	title = {Concrete Problems in AI Safety},
	author = {Amodei, Dario and others},
	journal = {arXiv preprint arXiv:1606.06565},
	year = {2016},
}

@misc{claudecode2025,
	title = {Claude Code},
	author = {{Anthropic}},
	howpublished = {\url{https://github.com/anthropics/claude-code}},
	year = {2025},
}

@misc{aws2025outage,
	title = {Summary of the {AWS} Service Event in the Northern Virginia {
	         (US-EAST-1)} Region},
	author = {{Amazon Web Services}},
	year = {2025},
}

@misc{azure2024outage,
	title = {Tracking the {Azure} Central {US} region outage},
	author = {{Microsoft}},
	year = {2024},
}

@misc{crowdstrike2024outage,
	title = {Falcon Sensor Content Update Preliminary Post Incident Report},
	author = {{CrowdStrike}},
	year = {2024},
}
	
\end{document}